\documentclass[runningheads]{llncs}

\usepackage{epsfig}
\begin{document}

\title{From Royal Road To Epistatic Road For Variable Length Evolution
Algorithm}
\titlerunning{From Royal Road To Epistatic Road}
\author{Michael Defoin Platel$^{1,2}$, Sebastien Verel$^1$, Manuel
Clergue$^1$ and
Philippe Collard$^1$}
\authorrunning{M. Defoin Platel et al.}
\institute{Laboratoire I3S, CNRS-Universit\'{e} de Nice Sophia Antipolis
\and ACRI-ST}

\maketitle

\begin{abstract} Although there are some real world applications where the
use of
variable length representation (VLR) in Evolutionary Algorithm is natural and
suitable, an academic framework is lacking for such representations. In
this work
we propose a family of tunable fitness landscapes based on VLR of
genotypes. The
fitness landscapes we propose possess a tunable degree of both neutrality and
epistasis; they are inspired, on the one hand by the Royal Road fitness
landscapes, and the other hand by the NK fitness landscapes. So these
landscapes
offer a scale of continuity from Royal Road functions, with neutrality and no
epistasis, to landscapes with a large amount of epistasis and no redundancy. To
gain insight into these fitness landscapes, we first use standard tools such as
adaptive walks and correlation length. Second, we evaluate the performances of
evolutionary algorithms on these landscapes for various values of the
neutral and
the epistatic parameters; the results allow us to correlate the
performances with
the expected degrees of neutrality and epistasis.
\end{abstract}

\section{Introduction} Individuals in Genetic Algorithms (GA) are generally
represented with strings of fixed length and  each position of the string
corresponds to one gene. So, the number of genes is fixed and each of them
can take
a fixed number of values (often $0$ and $1$). In variable length representation (VLR), like Messy GA or Genetic
Programming, genotypes have a variable number of genes.  Here, we consider VLR
where a genotype is a sequence of symbols drawn from a finite alphabet and
a gene
is a given sub-sequence of such symbols. The main difference with fixed length
representation is that a gene is identified by its form and not by its absolute
position in the genotype.

Some specific obstacles come with the variable length paradigm. One of the most
important is the identification of genes. Indeed, during recombination,
genes are
supposed to be exchanged with others that represent similar features. So the
question of the design of suitable crossover operators becomes essential
(see for
example \cite{DEFOIN:03}). Another difficulty due to variable length is the
tremendous amount of neutrality of the search space, as noted in
\cite{banzhaf:1997:emvsea}. Neutrality appears at different levels.  First,
a gene
may be located at different positions in the genotype. Second, some parts of genotype
(called introns) do not perform any functions and so do not contribute to
fitness. The last specificity is that variable length strings introduce a new
dimension in the search space,  which have to be carefully explored during
evolution to find regions where fitter individuals prevail.  The exploration of
sizes seems to be difficult to handle and may lead, as in Genetic
Programming,  to
an uncontrolled growth of individuals (a phenomenon called
bloat \cite{langdon97fitness}).

One of the major concerns in the GA field is to characterize the difficulty of
problems. One way to achieve this is to design problems with parameters
controlling
the main features of the search space; to run the algorithm; and to exhibit how
performances vary according to the parameters. With fixed length
representations,
some well known families exist, as the Royal Road functions, where inherent
neutrality is controlled by the block size, or the NK-landscapes, where the
tunable
parameter $K$ controls the ruggedness of the search space. With VLR, there
are only
a few attempts to design such academic frameworks\cite{daida99what}. Note, for
example, the Royal Tree \cite{PUNCH:96} and the Royal Road for Linear GP
\cite{DEFOIN:03}.

\section{Royal Road for variable length representation} In GA, Royal Road
landscapes (RR) were originally designed to describe how building blocks are
combined to produce fitter and fitter solutions and to investigate how the
schemata
evolution actually takes place \cite{FORREST&MITCHELL:93}. Little work is
related
to RR in variable length EA;   e.g. the Royal Tree Problem \cite{PUNCH:96}
which
is an attempt to develop a benchmark for Tree-based Genetic Programming
and which
has been used in Clergue et al. \cite{CLERGUE:02} to study problem difficulty.
To the best of our knowledge, there was no such work with linear structures.

In a previous work, we have proposed a new kind of fitness landscape \cite{DEFOIN:03},  called
Royal
Road landscapes for variable length EA (VLR Royal Road). Our aim was to
study the
behavior of a crossover operator during evolution.  To achieve this goal, we
needed experiments able to highlight the destructive (or constructive)
effects of
crossover  on building blocks.

To define VLR Royal Road, we have chosen a family of optimal genotypes and have
broken them into a set of small building blocks. Formally, the set of
optima is:
	\begin{displaymath}
\{g \in G_{\Sigma} \textrm{ }| \textrm{ } \forall l \in \Sigma  \textrm{, }
B_b(g,
l) = 1\},
	\end{displaymath} with
	\begin{displaymath} B_b(g, l) = \left \{
		\begin{array}{ll}	1		&  \textrm{if }
\exists\; i \in [0, \lambda-b] \textrm{ } |
							\textrm{ }, \forall
\;j \in [0,b-1] \textrm{, } g_{i+j} = l,	\\
					0		&
\textrm{otherwise},
		\end{array} \right.
	\end{displaymath} and
	\begin{itemize}
		\item
	$b \geq 1$ the size of blocks
		\item
	$\Sigma$ an alphabet of size $N$ that defines the set of all
possible letters $l$
per locus
		\item
	$G_{\Sigma}$ the finite set of all genotypes of size $\lambda \leq \lambda_{max}$\footnote{$\lambda_{max}$ have to be greater than $Nb$} defined over $\Sigma$
		\item
	$g$ a genotype of size $\lambda \leq \lambda_{max}$
		\item
	$g_{k}$ the $k^{th}$ locus of $g$.
	\end{itemize}
The following genotype $g \in G_{\Sigma}$ is an example of optimum,
with $\Sigma = \{A,T,G,C\}$ and $b=3$:
	\begin{displaymath}
g=\textbf{AAA}GTA\textbf{GGG}TAA\textbf{TTT}\textbf{CCC}TCCC\,.
	\end{displaymath}

$B_b(g, l)$ acts as a predicate accounting for the presence (or the
absence) of a
contiguous sequence of a single letter (i.e. a block). Note that only the
presence
of a block is taken into account, neither its position nor its repetition.
The {\it
number of blocks} corresponds to the number of letters $l \in \Sigma$  for
which
$B_b(g, l)$ is equal to one. In the previous example, only boldfaced sequences
contribute to fitness\footnote{Although the last sequence of 'CCC' is a valid block, it does not
contribute to fitness since it is only another occurrence.}. The contribution of each
block is fixed and so, the fitness $f_{Nb}(g)$ of genotype $g \in G_{\Sigma}$ having $n$ blocks is
simply:
	\begin{displaymath}
		f_{Nb}(g) = \frac{1}{N} \sum_{i=1}^{N} B_b(g,l_{i}) = \frac{n}{N}
	\end{displaymath}

To efficiently reach an optimum, the EA system has to create and combine
blocks
without breaking existing structures. These landscapes were designed in
such a way
that fitness degradation due to crossover may occur only when recombination
sites
are chosen inside blocks, and never in case of blocks translocations or
concatenations. In other words, there is no inter blocks epistasis.

\section{NK-Landscapes} Kauffman \cite{KAU:93} designed a family of
problems, the
NK-landscapes, to explore how epistasis is linked to the `ruggedness' of search
spaces. Here, epistasis corresponds to the degree of interaction between
genes, and
ruggedness is related to local optima, their number and especially their
density.
In NK-landscapes, epistasis can be tuned by a single parameter. Hereafter,
we give
a more formal definition of NK-landscapes followed by a summary review of their
properties.

\subsection{Definition} The fitness function of a NK-landscape is a
function $f_{NK}:
\lbrace 0, 1 \rbrace^{N} \rightarrow [0,1)$ defined on binary strings with $N$
loci. Each
locus $i$ represents a gene with two possible alleles, $0$ or $1$. An
'atom' with
fixed epistasis level is represented by a fitness components $f_i: \lbrace 0, 1
\rbrace^{K+1} \rightarrow [0,1)$ associated to each locus $i$. It depends
on the
allele at locus $i$ and also on the alleles at $K$ other epistatic loci
($K$ must
fall between $0$ and $N - 1$). The fitness $f_{NK}(x)$ of $x \in \lbrace 0, 1 \rbrace^{N}$
is the average of
the values of the $N$ fitness components $f_i$:\label{defNK}

$$ f_{NK}(x) = \frac{1}{N} \sum_{i=1}^{N} f_i(x_i; x_{i_1}, \ldots, x_{i_K})
$$ where $\lbrace i_1, \ldots, i_{K} \rbrace \subset \lbrace 1, \ldots, i -
1, i +
1, \ldots, N \rbrace$. Many ways have been proposed to choose the $K$ other
loci
from $N$ loci in the genotype. Two possibilities are mainly used: adjacent and
random neighborhoods. With an adjacent neighborhood, the $K$ genes
nearest to the
locus $i$ are chosen (the genotype is taken to have periodic boundaries).
With a
random neighborhood, the $K$ genes are chosen randomly on the genotype.
Each fitness component $f_i$ is specified by extension, ie a number $y_{i,(x_i; x_{i_1}, \ldots, x_{i_K})}$ from $[0, 1)$ 
is associated with each element 
$(x_i; x_{i_1}, \ldots, x_{i_K})$ from $\lbrace 0, 1 \rbrace^{K+1}$.
Those numbers are uniformly distributed in the interval $[0, 1)$.

\subsection{Properties} The NK-landscapes have been used to study links between
epistasis and local optima. The definition of local optimum is relative to a
distance metric or to a neighborhood choice. Here we consider that two strings of length $N$ are neighbors if their Hamming
distance is exactly one.
A string is a local optimum if it is fitter than its neighbors.\\
 The properties of
NK-landscapes are given hereafter in term of local optima: their
distribution of
fitness, their number and their mutual distance. These results can be found in
Kauffman\cite{KAU:93}, Weinberger\cite{WEI:91}, Fontana {\it et
al.}\cite{Fontana93RNA}.

\begin{itemize}
\item For $K = 0$ the fitness function becomes the classical additive
multi-locus
model, for which
\begin{itemize}
\item There is single and attractive global optimum.
\item There always exists a fitter neighbor (except for global optimum).
\item Therefore the global optimum could be reach on average in $N/2$ adaptive
steps.
\end{itemize}

\item For $K = N-1$, the fitness function is equivalent to a random
assignment of
fitnesses over the genotypic space, and so:
\begin{itemize}
\item The probability that a genotype is a local optimum is $\frac{1}{N+1}$.
\item The expected number of local optima is $\frac{2^N}{N+1}$.
\item The average distance between local optima is approximately $2 ln(N - 1)$
\end{itemize}
\label{propNK}
\item For $K$ small, the highest local optima share many of their alleles
in common.
\item For $K$ large:
\begin{itemize}
\item The fitnesses of local optima are distributed with an asymptotically
normal
distribution with mean $m$ and variance $s$ approximately:
$$
\begin{array}{rclcrcl} m & = & \mu + \sigma \sqrt{\frac{2 ln(K+1)}{K+1}} &
, & s &
= & \frac{ (K+1) \sigma^2 }{ N(K+1 + 2(K+2)ln(K+1))}
\end{array}
$$ where $\mu$ is the expected value of $f_i$ and $\sigma^2$ its variance.
In the
case of the uniform distribution, $\mu=1/2$ and $\sigma = \sqrt{1/12}$.
\item The average distance between local optima is approximately
$\frac{N log_2(K+1)}{2(K+1)}$.
\item The autocorrelation function $\rho(s)$ and the correlation length $\tau$
are:
$$
\begin{array}{rclcrcl}
\rho(s) & = & \left( 1 - \frac{ K+1 }{ N } \right)^s & , &
\tau    & = & \frac{ - 1 }{ ln( 1 - \frac{K+1}{N} ) }
\end{array}
$$.
\end{itemize}

\end{itemize}

\section{Epistatic Road for variable length representation} In this section, we
define a problem with tunable difficulty for variable length EA, called
Epistatic
Road functions (ER). To do so, we propose to use the relation between
epistasis and
difficulty.

	\subsection{Definition} Individuals in a variable length
representation may be
viewed as sets of interacting genes.  So, in order to model such a variable
length
search space, we have to first identify genes and second  explicitly define
their
relations. This can be easily done by extending the VLR Royal Road thanks to
dependencies between  the fitness contributions of blocks. Thus, genes are
designated as blocks and the contribution of a gene depends on the presence of
others,  exactly as in NK-landscapes.

More formally, the fitness function of an ER-landscape is a function
$f_{NKb} : G_{\Sigma} \rightarrow [0,1)$ defined on variable length genotypes. The fitness
components $f_i$ are defined in \ref{defNK}, and the fitness $f_{NKb}(g)$ of genotype $g \in G_{\Sigma}$ is
the average of $N$ fitness components $f_i$:
	\begin{displaymath}
		f_{NKb}(g) = \frac{1}{N} \sum_{i=1}^{N} f_i(B_b(g,l_{i});
B_b(g,l_{i_1}), \ldots,
B_b(g,l_{i_K}))
	\end{displaymath} 
In practice, we use an implementation of NK-landscape with
random neighborhood to compute $f_i$. We have to ensure that the set of all
genotypes having $N$ blocks corresponds to the end of the Road. For that
purpose,
first we exhaustively explore the space $\lbrace 0, 1 \rbrace^{N}$ to find  the
optimum value of the NK, then we permute this space in such a way that the
optimum
becomes $1^N$.

	\subsection{Tunability} The properties of an ER-landscape depends
on the three
parameters $N$, $K$  and $b$. Although these parameters are not entirely
independent,  each allows us to control a particular aspect of the landscape.
Increasing the parameter $N$ causes the size of both the search space and the
neighborhood of genotype to increase. Moreover, as $N$ determines the
number of
genes to find,  the computational effort required to reach the optimum will
be more
important when high values of $N$ are used. The parameter $b$ controls the
degree
of neutrality. As $b$ increases the size of iso-fitness sets increases.
Finally,
the parameter $K$ allows to control the number of epistatic links between
genes
and so the number of local optima. For $K=0$, an ER-landscape will be very
closed
to the corresponding VLR Royal Road  since insertion of a new block in a
genotype
always increases the fitness. In contrast, for $K=N-1$, with a high level of
epistasis, the vast majority of the roads leads to local optima where the
insertion of a
new block in a genotype always decreases the fitness.

\section{Fitness landscape analysis} Many measures have been developed to
describe
fitness landscapes in terms of ``difficulty''. Here, ``difficulty'' refers
to the
ability of a local heuristic to reach the optimum. In this section some of
those
metrics are applied to the ER-landscapes. In particular, we show how difficulty
changes according to the three parameters $N$, $K$ and $b$. The
neighborhood of
variable length genotypes is different from the neighborhood of fixed length
genotypes.  To define a neighborhood in ER-landscapes, we use String Edit
Distance, like \textit{Levenshtein distance}
\cite{LEVENSHTEIN:66} which has been already used in GP to compute or control
diversity \cite{BRAMEIER&BANZHAF:01},  or to study the influence of genetic
operators \cite{OREILLY:97}. By definition, the Edit Distance between two
genotypes
corresponds to the minimal number of elementary operations  (deletion,
insertion
and substitution) required to change one genotype into the other. So two
strings in
the search space are neighbors if the Edit Distance between them is equal
to $1$.
Thus a string of length $\lambda$ has $(2 \lambda + 1)N$ neighbors.

In order to minimize the influence of the random creation of an
NK-landscape, we
take the average of the following measures over $10$ different landscapes
for each
couple of parameters $N$ and $K$.  We have perform experiments for $N=8$,
$10$ and
$16$, for $K$ between $0$ and $N - 1$ and for $b$ between $1$ and $5$.

\subsection{Random walks, autocorrelation function and correlation length}
Weinberger\cite{WEI:91,WEI:90} defined {\it autocorrelation function} and {\it
correlation length} of random walks to measure the epistasis of fitness
landscapes.\\
A random walk $\lbrace g_t, g_{t+1}, \ldots \rbrace$ is a series where
$g_t$ is
the initial genotype and $g_{i+1}$ is a randomly selected neighbor
of $g_i$.
Then the autocorrelation function $\rho$ of a fitness function $f$ is the autocorrelation function of the
time series $\lbrace f(g_t), f(g_{t+1}), \ldots \rbrace\,$:
$$
\rho(s) = \frac{\langle f(g_t) f(g_{t+s}) \rangle_{t} - \langle f
\rangle^2}{var(f)}
$$
The correlation length $\tau$ measures how the correlation
function decreases and so how rugged the landscape is. More rugged the
landscape
the shorter the correlation length.
$$\tau = - \frac{1}{ln(\rho(1))}$$

Empirical measures on ER landscapes were performed on $20.10^3$ random walks of
length $35$ for each triplet of parameters $N$, $K$, $b$ and for each of $10$
instances of NK-landscapes.  The initial genotypes were generated by randomly
choosing its length between $0$ and $\lambda_{max}$ and  then randomly choosing
each letter of the genotype. For those random walks, $\lambda_{max}$ is
equal to $2 N b$.
	\begin{figure}[!htbp]
		\begin{center}
		\begin{minipage}[t]{.46\linewidth}
			\includegraphics[width=160pt,
height=140pt]{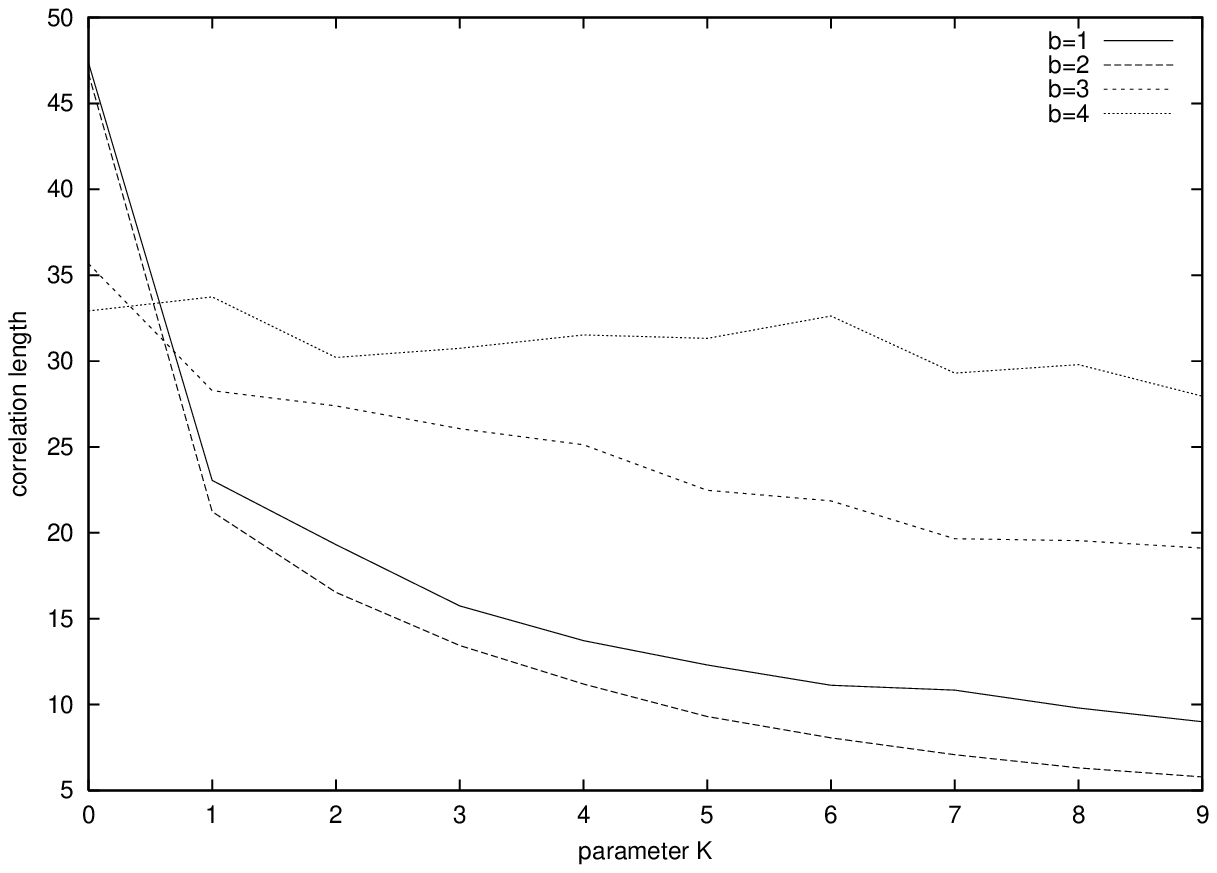}
			\caption{Mean correlation length of ER-landscapes
for $N=10$}
			\label{fig_corLen}
		\end{minipage} \hfill
		\begin{minipage}[t]{.46\linewidth}
			\includegraphics[width=160pt,
height=140pt]{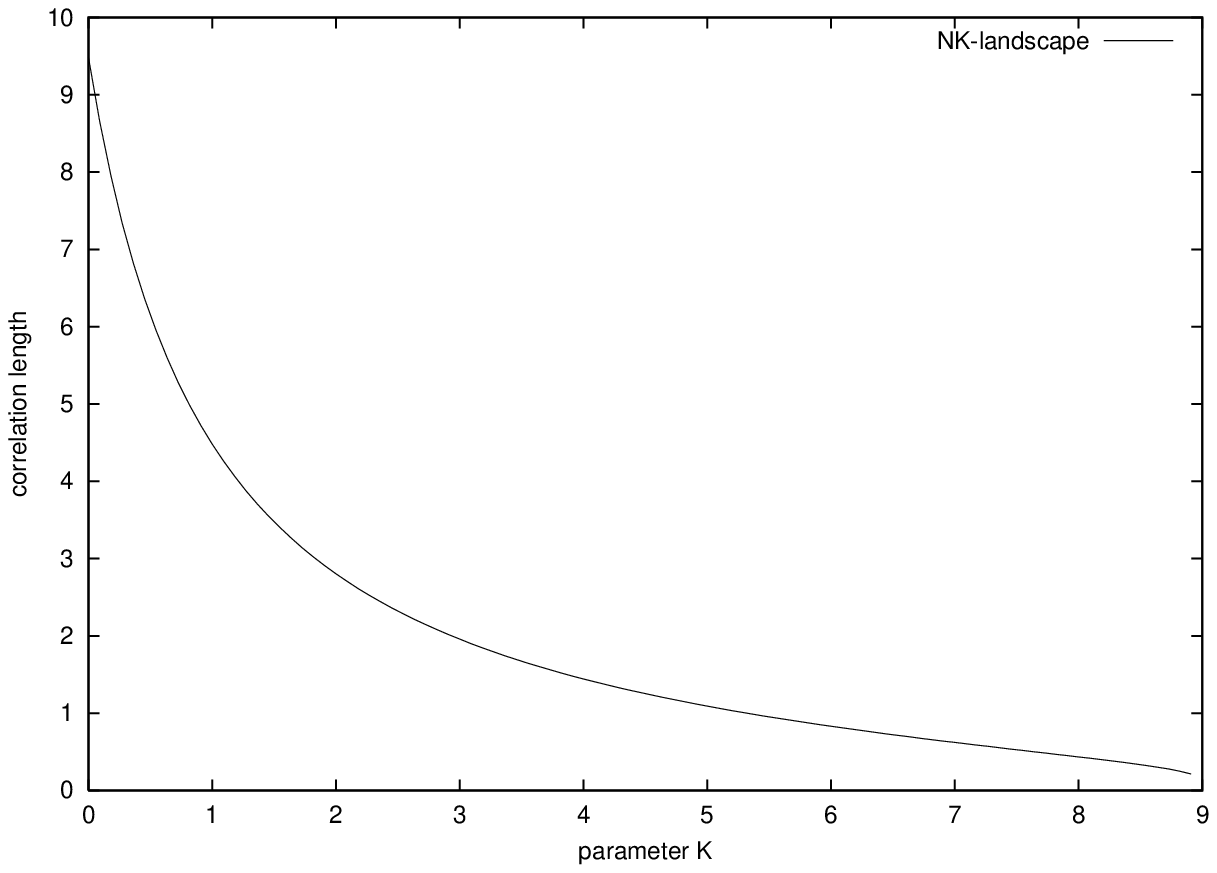}
			\caption{Theoretical correlation length of
NK-landscapes for $N=10$}

			\label{fig_corLenNK}
		\end{minipage}
		\end{center}
	\end{figure}
	\begin{figure}[!htbp]
		\begin{center}
		\begin{minipage}[t]{.46\linewidth}
			\includegraphics[width=160pt,
height=140pt]{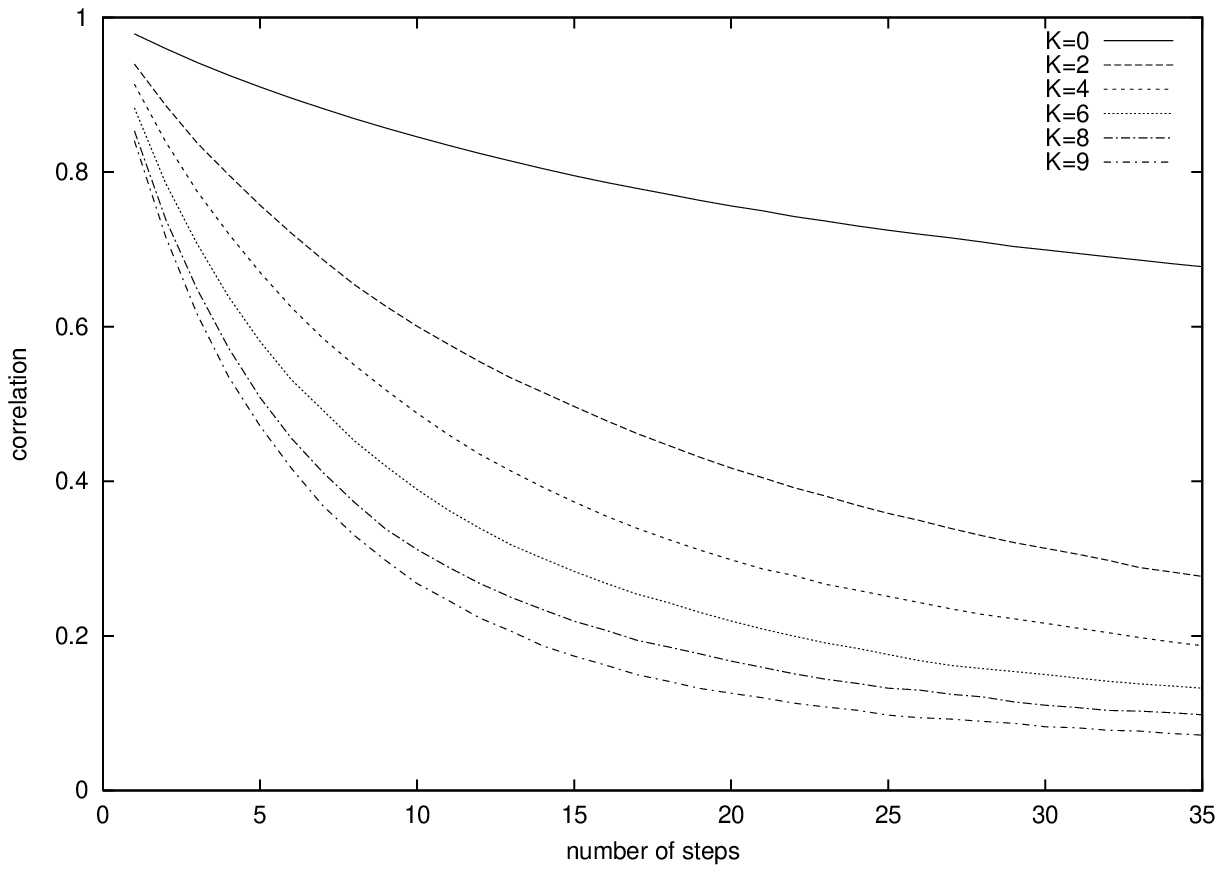}
			\caption{Autocorrelation function of ER-landscape
for $N=10$}
			\label{fig_corFunc}
		\end{minipage} \hfill
		\begin{minipage}[t]{.46\linewidth}
			\includegraphics[width=160pt, height=140pt]{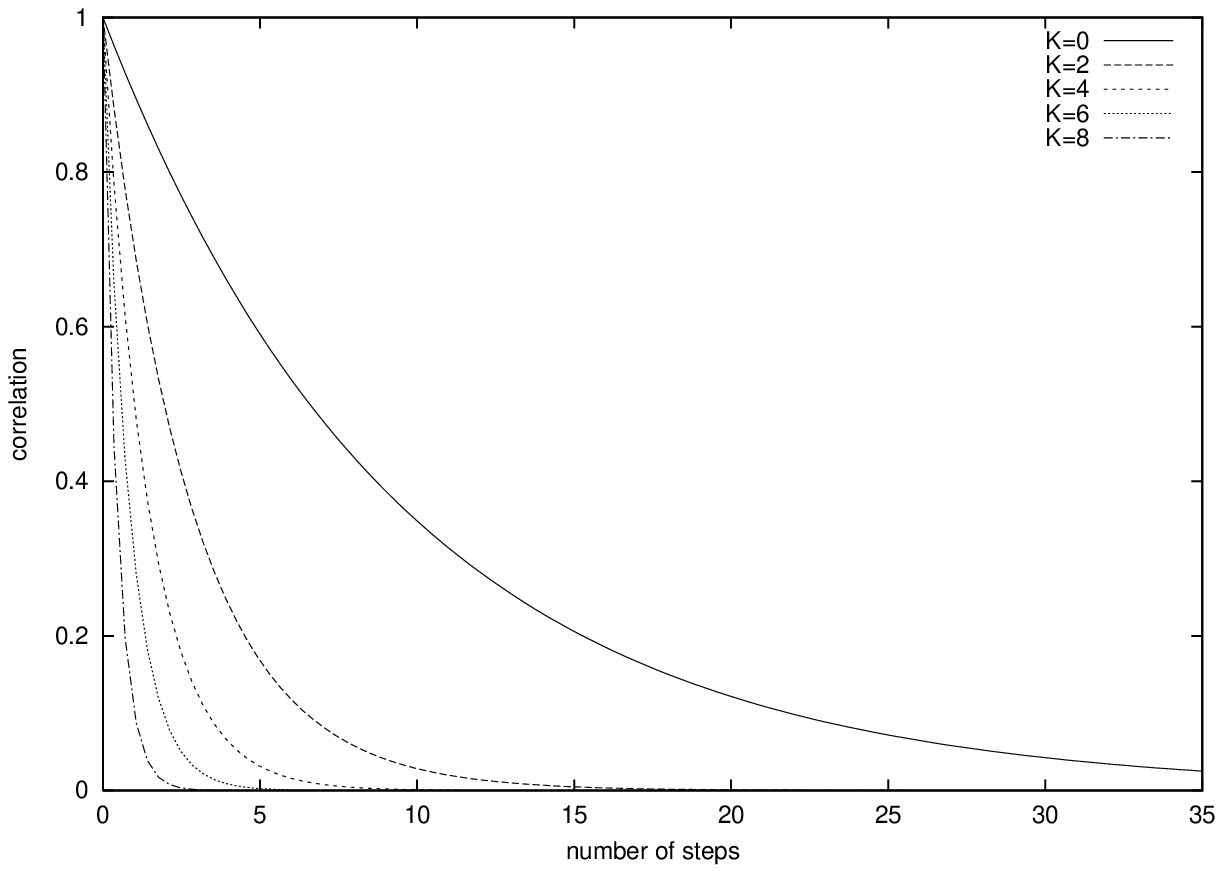}
			\caption{Theoretical autocorrelation function of
NK-landscape for $N=10$}
			\label{fig_corFuncNK}
		\end{minipage}
		\end{center}
	\end{figure} For small values of $b$, the correlation length
decreases quickly
(when the parameter $K$ increases, see fig. \ref{fig_corLen} and
\ref{fig_corFunc}). As expected, the correlation of fitness between genotypes
decreases with the modality due to the parameter $K$. We can compare this
variation
with the theoretical correlation length of NK-landscapes, given in \ref{propNK} (see fig.
\ref{fig_corLenNK}
and
\ref{fig_corFuncNK}). As $b$ increases, the influence of $K$ on the correlation
length decreases. Neutrality keeps a high level of correlation in spite of the
increase in modality.
\subsection{Adaptive walks and local optima} Several variants of adaptive walk 
(often called myopic or greedy adaptive walk) exists. Here we use the series 
$\lbrace g_t, g_{t+1}, \ldots , g_{t+l} \rbrace$ where $g_t$ is the
initial genotype and $g_{i+1}$ is one of the fittest neighbor of $g_i$.
The walk stops on $g_{t+l}$ which is a local optimum. By computing several adaptive
walks, we can estimate:%
\begin{itemize}
\item The fitness distribution of local optima by the distribution of the final
fitnesses $f(g_{t+l})$.
\item The distance between local optima which is approximately twice the
mean of
the length $l$ of those adaptive walks.
\end{itemize}

Empirical measurements on ER landscapes were performed on $2.10^3$ random
walks for
each triplet of parameters $N$, $K$, $b$ and for each of $10$ instances of
NK-landscapes. We used the same initialization procedure as the random
walk. The parameter $\lambda_{max}$ is set to $50$.
	\begin{figure}[!htbp]
		\begin{center}
		\begin{minipage}[t]{.46\linewidth}
			\includegraphics[width=160pt,
height=140pt]{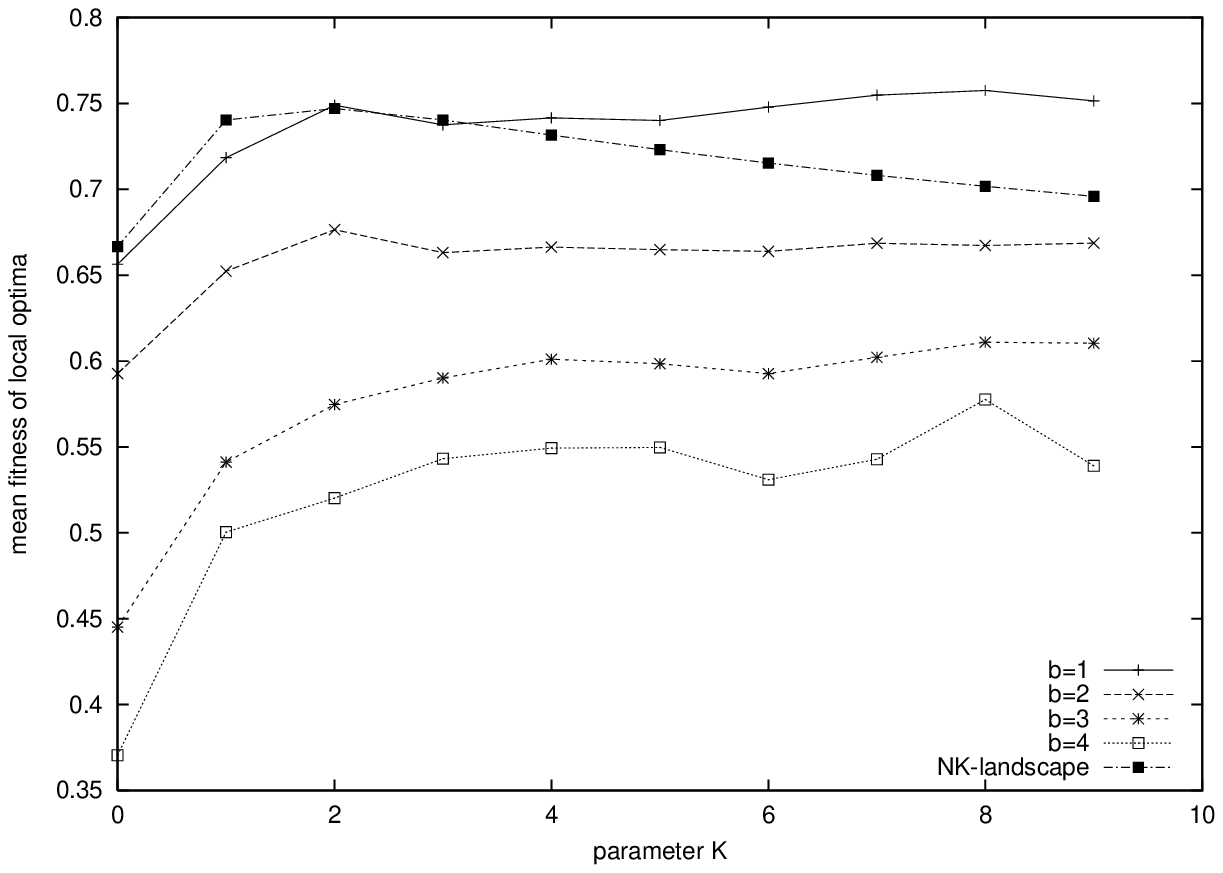}
			\caption{Mean fitness of local optima of
ER-landscapes obtained with adaptive
walks for $N=10$}
			\label{fig_meanFit}
		\end{minipage} \hfill
		\begin{minipage}[t]{.46\linewidth}
			\includegraphics[width=160pt,
height=140pt]{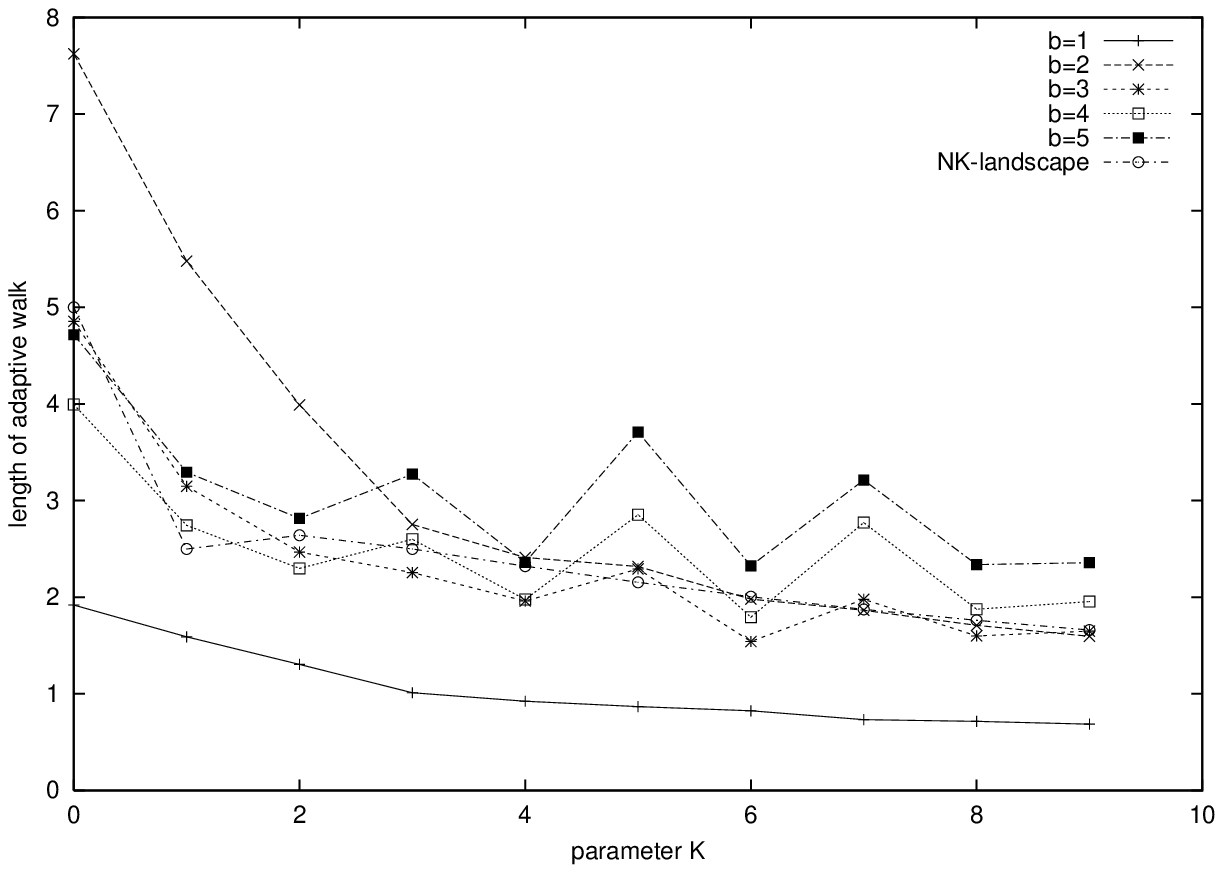}
			\caption{Mean length of adaptive walks on
ER-landscape for $N=10$}

			\label{fig_meanLen}
		\end{minipage}
		\end{center}
	\end{figure}%
 The distribution of local optima fitnesses is close to
normal
distribution. The mean fitness of local optima is represented for $N=10$ on
Figure \ref{fig_meanFit}; it decreases with $b$. The variations of the
fitness of
local optima are great for small values of $K$ but become almost
insignificant for
medium and high values of $K$. In Figure \ref{fig_meanLen}, the mean length of
adaptive walks is represented for $N=10$. As expected, it decreases with
$K$ for
small values of $b$.  So, the parameter $K$ increases the ruggedness of the
ER-landscape.  On the other hand, when $b$ is higher, $K$ has less
influence on the length of the walk. Indeed, the adaptive walk breaks off
more often on neutral plateaux.

\subsection{Neutrality} A random walk is used to measure the neutrality of
ER-landscapes.  At each step, the number of neighbors with lower, equal
and higher
fitness is counted.  We perform $2.10^3$ random walks of length $20$ for each
triplet of parameter $N$, $K$ and $b$.  The Table \ref{tab_neutre} gives the
proportions of such neighbors  for $N$=$8$, $K$=$4$ (they depend slightly
on $N$
and $K$) and for several values of $b$.  The number of equally fit
neighbors is
always high and is maximum for $b$=$4$.  So, neutral moves are a very important
feature of ER-landscapes.
	\begin{table}[!htbp]
		\begin{center}
		\caption{Proportion of Lower, Equal and Higher neighbor}
		\label{tab_neutre}
		\begin{tabular}{|c| c | c | c |}
		\hline
		& \multicolumn{3}{c|}{$N=8$, $K=4$ }	\\
		\cline{2-4}
		Block size	& Lower	& Equal		& Higher\\
		\hline
		\hline
		$b = 2$		& 7.2	&  85.8		&  7.0	\\
		$b = 3$		& 2.8	&  94.4		&  2.8	\\
		$b = 4$		& 0.5	&  98.9		&  0.6	\\
		\hline
       		\end{tabular}
		\end{center}
	\end{table}
\section{EA performances} In this section, we want to compare the
performances of
an evolutionary system on ER-landscapes for various settings of the three parameters $N$,
$K$ and
$b$. The performances are measured by the success rate
and the
mean number of blocks found. In order to minimize the influence of the random
creation of NK-landscapes, we take the average of these two measures over $10$
different landscapes.
$35$ independent runs are performed with mutation and crossover rates of
respectively $0.9$ and $0.3$ (as found in \cite{DEFOIN:03}). The standard
one point
crossover, which blindly swaps sub-sequences of parents, was used. Let us
notice
that a mutation rate of $0.9$ means that each program involved in
reproduction  has
a $0.9$ probability to undergo one insertion, one deletion and one
substitution.
Populations of $1000$ individuals were randomly created according  to a maximum
creation size of $50$. The evolution, with elitism, maximum program size of
$100$($\lambda_{max}$),
$4$-tournament selection,  and steady-state replacement, took place during
$400$
generations.

\subsection{Results} 
\begin{figure}[!htbp]
		\begin{center}
		\includegraphics[width=320pt, height=200pt]{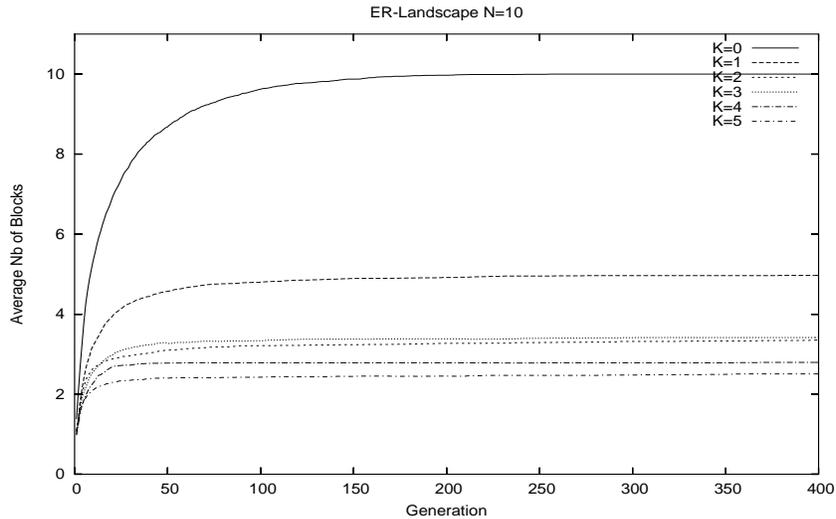}

		\caption{Evolution of average number of blocks found on ER
$N$=$10$ and $b$=$4$.}
		\label{fig_evol}
		\end{center}
	\end{figure}

\begin{figure}[!htbp]
		\begin{minipage}[t]{.46\linewidth}
		\begin{center}
		\includegraphics[width=160pt, height=140pt]{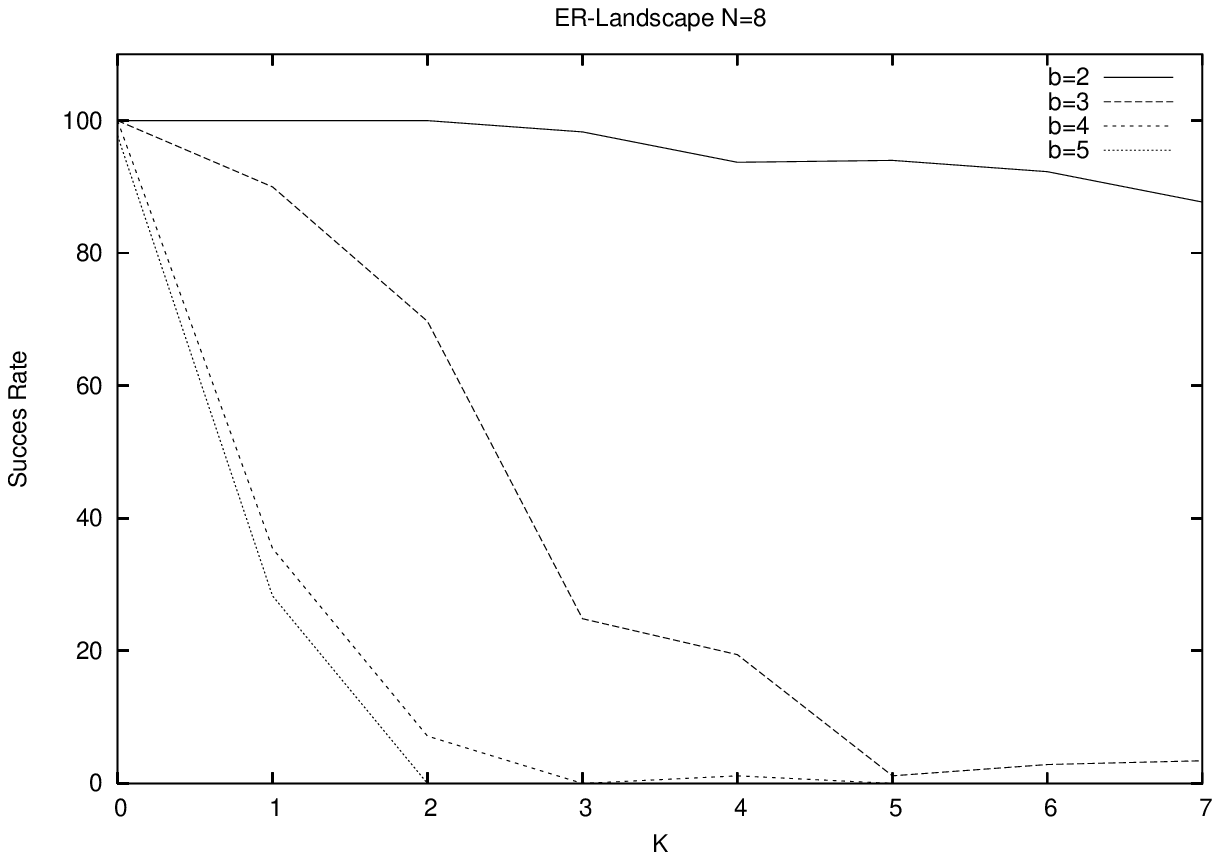}
			\caption{Success rate as a function of K on ER
$N$=$8$.}
		\label{fig_succes}
		\end{center}
		\end{minipage} \hfill
		\begin{minipage}[t]{.46\linewidth}
		\begin{center}
		\includegraphics[width=160pt, height=140pt]{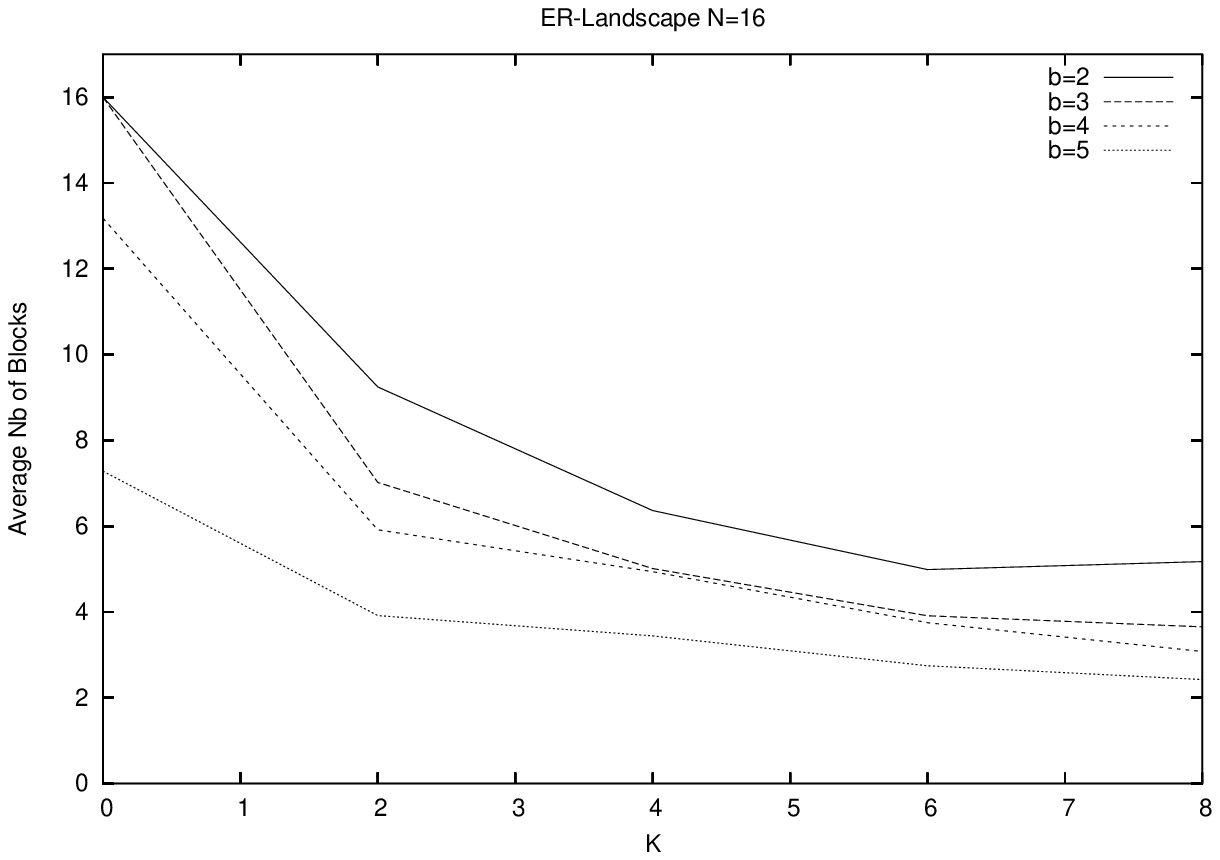}

		\caption{Average number of blocks found as a function of K
on ER $N$=$16$.}
		\label{fig_number}
		\end{center}
		\end{minipage}
	\end{figure}

We have performed experiments for $N$=$8$, $10$ and
$16$, for
$K$ between $0$ and $N/2$ and  for $b$ between $2$ and $5$.  We note that
the case
$b$=$1$ is not relevant because the optimum is always found at the first
generations for all values of $K$. In Figure \ref{fig_succes}, we have
reported the
success rate (over $35\times10$ runs) as a function of $K$ for $N$=$8$. As
expected, we see that for $K$=$0$, the problem is easy to solve for all
values of
$b$. Moreover, increasing $K$ decreases the success rate and this
phenomenon is
amplified when high values of $b$ are used. For $N$=$10$ and $16$, too few runs
find the optimum and so the variations of the success rate are not
significant. The
Figure \ref{fig_evol} gives the evolution of the average number of blocks
of the
best individual  found for $N$=10, $b$=4 and $K$ between $0$ and $5$.  At the
beginning of the runs, the number of blocks found increases quickly then halts
after several generations. The higher is $K$, the sooner ends evolution.  This
behavior looks like premature converge and confirms experimentally that
the number
of local optima increases with $K$. We have also plotted the average number of
blocks of the best individual found as a function of $K$ for $N$=$16$  (see
Fig.
\ref{fig_number}).  We see that this number decreases as $K$ or $b$ increases.
These two parameters undoubtedly modify the performances and  can be used
independently to increase problem difficulty.

In \cite{jr94fitness}, random and adaptive walks have been used to measure
problem
difficulty in GP. The author has shown that only the adaptive walk gives
significant results on classical GP benchmarks. We have computed the
correlation between these two measures and the average number of blocks
found on
ER,  for all settings of $N$, $K$ and $b$. We note that the correlation is
$0.71$
between the length of the adaptive walk and the number of blocks.
Conversely, the
length of the random walk seems to be completely uncorrelated to performance.

\section*{Conclusion} We think that a better understanding of the
implications of
variable length representations on Evolutionary Algorithms would allow
researchers
to use these structures more efficiently. In this paper, our goal is to
investigate
which kind of property could influence the difficulty of such problems. We have
chosen two features of search spaces, the neutrality and the ruggedness. So, we
have designed a family of problems, the Epistatic Road landscapes, where those
features can be tuned independently.

Statistical measures computed on ER-landscapes have shown that, similarly to
NK-landscapes,
tuning the epistatic coupling parameter $K$ increases ruggedness.
Moreover, as for Royal Roads functions, tuning the size block
parameter
$b$ increases neutrality.

The experiments that we have performed with a VLR evolutionary algorithm, have demonstrated the
expected difficulty according to parameters $b$ and $K$. Although our
results can
not be directly transposed to real world problems,  mainly because our initial
hypotheses are too simple, in particular about the nature of building blocks,
we have a
ready-to-use VLR problem of tunable difficulty, which allows us to study the
effects of genetic operators  and the dynamics of the evolutionary process.

\bibliographystyle{splncs}
\bibliography{Biblio}

\end{document}